\pdfoutput=1

\documentclass[11pt]{article}

\usepackage{ACL2023}

\usepackage{times}
\usepackage{latexsym}

\usepackage[T1]{fontenc}

\usepackage[utf8]{inputenc}

\usepackage{microtype}

\usepackage{inconsolata}

\usepackage{subcaption, booktabs}
\usepackage{graphicx}
\usepackage{xcolor}
\usepackage{colortbl}
\usepackage{bm}
\usepackage{amsmath, amssymb}
\usepackage{fdsymbol}
\usepackage{amsfonts}
\usepackage{multirow}
\usepackage{enumitem}
\usepackage{physics}
\usepackage{tabularx}

\usepackage{tikz}
\usetikzlibrary{fit}
\usetikzlibrary{shapes}

\usepackage{amsmath}
\DeclareMathOperator*{\argmin}{argmin}

\definecolor{my-color}{rgb}{0.28, 0.58, 0.28}
\definecolor{Gray}{gray}{0.85}

\title{Parameter-Efficient Fine-Tuning without Introducing New Latency}

\author{Baohao Liao, Yan Meng, Christof Monz \\
  Language Technology Lab, University of Amsterdam \\
  \texttt{\{\href{mailto:b.liao@uva.nl}{b.liao}, \href{mailto:y.meng@uva.nl}{y.meng}, \href{mailto:c.monz@uva.nl}{c.monz}\}@uva.nl}
  }

\begin{document}
\maketitle
\begin{abstract}
Parameter-efficient fine-tuning (PEFT) of pre-trained language models has recently demonstrated remarkable achievements, effectively matching the performance of full fine-tuning while utilizing significantly fewer trainable parameters, and consequently addressing the storage and communication constraints. Nonetheless, various PEFT methods are limited by their inherent characteristics. In the case of sparse fine-tuning, which involves modifying only a small subset of the existing parameters, the selection of fine-tuned parameters is task- and domain-specific, making it unsuitable for federated learning. On the other hand, PEFT methods with adding new parameters typically introduce additional inference latency. In this paper, we demonstrate the feasibility of generating a sparse mask in a task-agnostic manner, wherein all downstream tasks share a common mask. Our approach, which relies solely on the magnitude information of pre-trained parameters, surpasses existing methodologies by a significant margin when evaluated on the GLUE benchmark. Additionally, we introduce a novel adapter technique that directly applies the adapter to pre-trained parameters instead of the hidden representation, thereby achieving identical inference speed to that of full fine-tuning. Through extensive experiments, our proposed method attains a new state-of-the-art outcome in terms of both performance and storage efficiency, storing only 0.03\% parameters of full fine-tuning.\footnote{Code at \href{https://github.com/baohaoliao/pafi_hiwi}{https://github.com/baohaoliao/pafi\_hiwi}} 

\end{abstract}

\begin{figure}[t]
  \centering
    \includegraphics[width=0.45\textwidth]{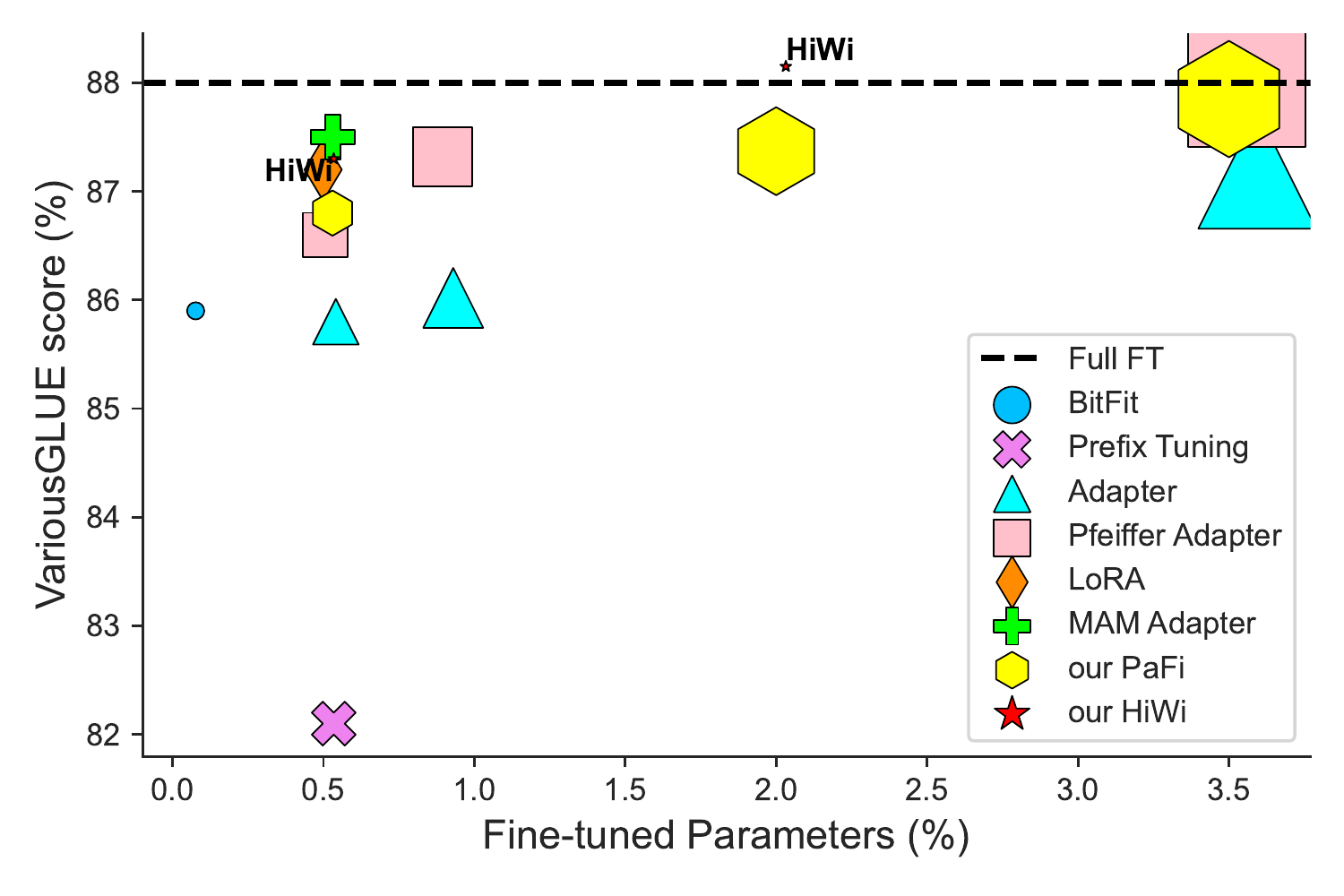}
    \caption{Comparison among various PEFT methods. (Zoom in for a better experience.) Marker size denotes the storage requirement. HiWi stores the least and its storage is invariant to the number of trainable parameters. Specific values are shown in Table \ref{tab: all}.}
    \label{fig: overall}
\end{figure}

\section{Introduction}
Pre-trained language models (PLMs) have served as a cornerstone for various natural language processing applications, favoring downstream tasks by offering a robust initialization \cite{DBLP:conf/naacl/PetersNIGCLZ18, DBLP:conf/naacl/DevlinCLT19, DBLP:journals/corr/abs-1907-11692, DBLP:journals/tacl/LiuGGLEGLZ20, DBLP:conf/nips/BrownMRSKDNSSAA20, DBLP:conf/emnlp/LiaoTHNM22}. 
Starting with a pre-trained checkpoint, a model can achieve significantly better performance on tasks of interest than the one from scratch. 
The most historically common way to adapt PLMs to downstream tasks is to update all pre-trained parameters, \textit{full fine-tuning}. 
While full fine-tuning produces numerous state-of-the-art results, it is impractical for storage-constrained and communication-frequent cases, like federated learning \cite{DBLP:conf/aistats/McMahanMRHA17}, since it requires a full copy of the fine-tuned model for each task. 
This issue becomes more severe when PLMs are large-scale \cite{DBLP:conf/nips/BrownMRSKDNSSAA20, DBLP:journals/corr/abs-2205-01068, DBLP:journals/corr/abs-2203-15556, DBLP:journals/jmlr/RaffelSRLNMZLL20, DBLP:journals/corr/abs-2211-05100, DBLP:journals/corr/abs-2302-13971}, the number of tasks in interest grows, or data are privately saved on hundreds of servers for federated learning.

An alternative approach popularized by \citet{DBLP:conf/icml/HoulsbyGJMLGAG19} is \textit{parameter-efficient fine-tuning} (PEFT), where a small number of task-specific parameters is updated and the majority of PLM's parameters is frozen.
In this way, only one general PLM alongside the modified parameters for each task is saved or transferred. Except for saving memory and training cost, PEFT matches the performance of full fine-tuning with only updating less than $1\%$ of the PLM parameters, quickly adapts to new tasks without catastrophic forgetting \cite{DBLP:conf/eacl/PfeifferKRCG21} and often exhibits robustness in out-of-distribution evaluation \cite{DBLP:conf/acl/LiL20}. These compelling advantages have sparked considerable interest in the adoption of PEFT.

PEFT methods can be split into two categories: sparse and infused fine-tuning. Sparse fine-tuning tunes a small subset of existing parameters without introducing new parameters. One typical example is BitFit \cite{DBLP:conf/acl/ZakenGR22}, where only the biases are updated. Nonetheless, BitFit is not scalable because of the fixed bias terms. Diff Pruning \cite{DBLP:conf/acl/GuoRK20} and FISH Mask \cite{DBLP:conf/nips/SungNR21} alleviate this issue by learning and updating task-specific masked parameters with a specified sparsity ratio. However, different masks are learned under different tasks, making these two methods unsuitable for the federated learning setting, where data is rarely i.i.d. across servers.

Infused fine-tuning introduces new parameters to PLMs, and only updates these parameters during training. For example, adapter fine-tuning \cite{DBLP:conf/icml/HoulsbyGJMLGAG19, DBLP:conf/eacl/PfeifferKRCG21} inserts adapters to each layer of the PLM. Other methods, like Prefix Tuning \cite{DBLP:conf/acl/LiL20} and Prompt Tuning \cite{DBLP:conf/emnlp/LesterAC21}, append trainable vectors to input or hidden layers. However, inference latency is typically introduced by the newly added parameters and is nonnegligible for some complex tasks, like machine translation (MT) and summarization that add more than $4\%$ of the PLM parameters \cite{DBLP:conf/iclr/HeZMBN22}.

\begin{figure*}[t]
  \centering
    \includegraphics[width=0.9\textwidth]{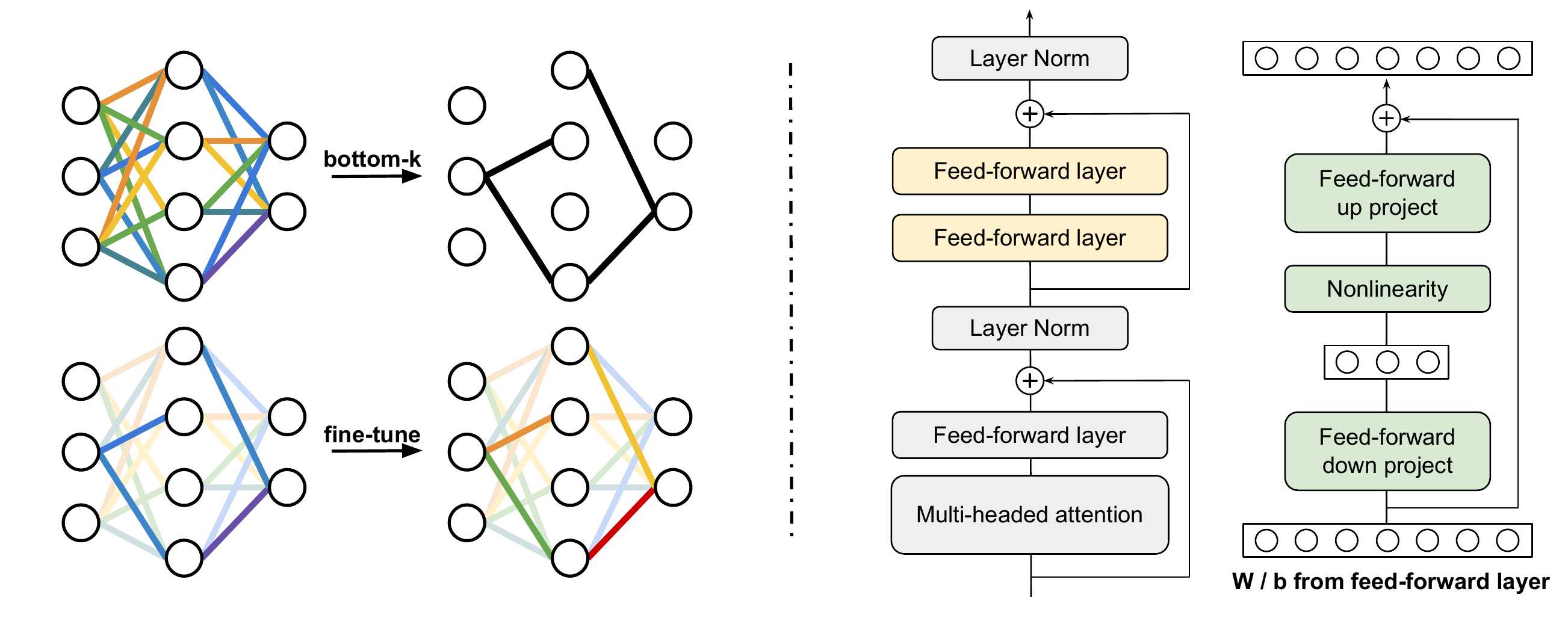}
    \caption{Our proposed methods. \textbf{PaFi} (left): We specify the pre-trained parameters with the smallest absolute magnitude as trainable (upper plot) and only update them during fine-tuning (bottom plot). \textbf{HiWi} (right): Instead of feeding hidden representation to an adapter, we input the pre-trained weights or biases from the feed-forward layers (in yellow) to the adapter and throw the adapter away after training. Only the modules in green are trainable.}
    \label{fig: pafi hiwi arch}
\end{figure*}

In this paper, we address the above-mentioned challenges from sparse and infused fine-tuning by proposing two methods, PaFi and HiWi (illustrated in Figure \ref{fig: pafi hiwi arch}). PaFi is a sparse fine-tuning method that selects trainable parameters in a task-agnostic way. I.e., we have the same mask for various downstream tasks. The mask generation of PaFi is also data-less. It doesn't require any training on any data. HiWi is an infused fine-tuning method that applies the adapters directly to pre-trained weights or biases instead of to hidden representations. After training, the adapters are abandoned, therefore sharing the same inference speed as full fine-tuning.

Our main contributions in this paper are: (1) We introduce two novel transfer learning methods that solve the above-mentioned key challenges of sparse and infused fine-tuning. (2) We empirically evaluate PaFi on the GLUE benchmark and show its effectiveness over existing sparse fine-tuning methods. (3) We compare our methods to a wide range of baselines on a newly constructed benchmark that contains tasks in different types and resources. HiWi outperforms all baselines and full fine-tuning, while requiring the minimum storage (see Figure \ref{fig: overall}). (4) Our proposed methods still show their effectiveness on a complex task, i.e. machine translation. And all PaFi and HiWi share the same inference speed as full fine-tuning.

\section{Preliminaries}
\label{sec: preliminaries}
In this section, we give an overview of sparse fine-tuning and adapter fine-tuning, and highlight the key challenges of these PEFT methods. 

\textbf{Sparse Fine-Tuning}. Sparse fine-tuning formulates a task-specific fine-tuning as a two-phase learning problem. In the first phase, one needs to determine which subset of the pre-trained parameters $\bm{\theta}^{(0)}$ can be modified by generating a sparse mask $\bm{m} \in \{0, 1\}^{|\bm{\theta}^{(0)}|}$, where 1s in $\bm{m}$ denote the corresponding parameters are trainable. BitFit \cite{DBLP:conf/acl/ZakenGR22} heuristically specifies the bias terms trainable. Diff Pruning \cite{DBLP:conf/acl/GuoRK20} and LT-SFT \cite{DBLP:conf/acl/AnsellPKV22} fully fine-tune PLM on downstream task to obtain $\bm{\theta}^{(1)}$. And the $k$ parameters with the greatest absolute difference $|\bm{\theta}^{(1)} - \bm{\theta}^{(0)}|$ are selected for updating in the next phase. FISH Mask \cite{DBLP:conf/nips/SungNR21} uses the gradient information of $\bm{\theta}^{(1)}$ to learn the mask.

After obtaining the mask, the PLM is fine-tuned and only the masked parameters are updated whereas the others are frozen. The learning procedure of the second phase is defined as
\begin{align}
    \bm{\theta}^{(2)} = \argmin_{\bm{\theta}} \mathcal{L}(\mathcal{D}; \bm{m} \odot \bm{\theta})
\label{eq: sparse fine-tuning}
\end{align}
with $\bm{\theta}$ initialized by $\bm{\theta}^{(0)}$, where $\mathcal{L}$ and $\mathcal{D}$ are the objective and data of downstream task, respectively. In addition, $(\bm{1}-\bm{m}) \odot \bm{\theta}^{(2)}=(\bm{1}-\bm{m}) \odot \bm{\theta}^{(0)}$, since we only update the masked parameters. In the end, only a common $\bm{\theta}^{(0)}$, the updated parameters $\bm{m} \odot \bm{\theta}^{(2)}$ and their indices are saved, which is storage-friendly with a large number of downstream tasks.

\textbf{Adapter Fine-Tuning}. Adapter fine-tuning methods \cite{DBLP:conf/icml/HoulsbyGJMLGAG19, DBLP:conf/eacl/PfeifferKRCG21} insert one or multiple small MLP modules into each layer of the PLM. This MLP module consists of a down ($\bm{W}_{down} \in \mathbb{R}^{d \times r}$) and up ($\bm{W}_{up} \in \mathbb{R}^{r \times d}$) projection pair, where $r$ is the bottleneck dimension, $d$ is the dimension of hidden representation and $r \ll d$. Most adapter fine-tuning methods can be fed into the formula of
\begin{align}
    \bm{h} \leftarrow \bm{h} + f(\bm{h} \bm{W}_{down})\bm{W}_{up}
\label{eq: normal adapter}
\end{align}
where $\bm{h}$ is the input to the adapter and $f(\cdot)$ is a nonlinear function.

The adapter fine-tuning methods in the formula of Equation \ref{eq: normal adapter} are module-wise, which means they consider the attention module or the feed-forward module as a unit and insert the adapter in between or after these units. In contrast, LoRA \cite{DBLP:conf/iclr/HuSWALWWC22} inserts the adapter layer-wise as:
\begin{align}
    \bm{h} \leftarrow \bm{h}\bm{W} + \bm{h}\bm{W}_{down}\bm{W}_{up}
\label{eq: lora}
\end{align}
where $\bm{W} \in \mathbb{R}^{d \times d}$ is a pre-trained weight. LoRA has the same inference speed as full fine-tuning, since we can pre-compute $\bm{W} \leftarrow \bm{W} + \bm{W}_{down}\bm{W}_{up}$ and use the new $\bm{W}$ for inference. Both sparse fine-tuning and adapter fine-tuning can be fed into a unified framework.

\textbf{A Unified Framework for Sparse and Adapter Fine-Tuning}.
Normally, we initialize $\bm{W}_{down}$ and $\bm{W}_{up}$ (or at least one of them) close to $\bm{0}$ \cite{DBLP:conf/icml/HoulsbyGJMLGAG19, DBLP:conf/iclr/HuSWALWWC22}, so the initial state of $\bm{h}$ is close to the original state $\bm{h}$ of PLM, which makes the fine-tuning empirically perform better. This initialization is important for PEFT in case of the catastrophic forgetting of the pre-training knowledge. Supposed we defined the newly added parameters ($\bm{W}_{down}$s and $\bm{W}_{up}$s) for adapter fine-tuning as $\bm{\delta}$, the initialization of $\bm{\delta}$ (i.e. $\bm{\delta}^{(0)} \approx \bm{0}$) fulfills $\mathcal{L}^{pre}(\cdot;\bm{\phi}^{(0)}) \approx \mathcal{L}^{pre}(\cdot; \bm{\theta}^{(0)})$, where $\mathcal{L}^{pre}$ is the pre-training objective\footnote{$\mathcal{L}^{pre}$ can't be replaced with $\mathcal{L}$ that is the objective for downstream task, since most downstream tasks require a randomly initialized classifier, which makes $\mathcal{L}$ unpredictable.} and $\bm{\phi}^{(0)} = \{\bm{\theta}^{(0)}, \bm{\delta}^{(0)}\}$. 

Straightforwardly, we can combine the second phase of sparse fine-tuning (Equation \ref{eq: sparse fine-tuning}) and the adapter fine-tuning as:
\begin{align}
    \hat{\bm{\phi}} = \argmin_{\bm{\phi}} \mathcal{L}(\mathcal{D}; \bm{z} \odot \bm{\phi})
\label{eq: unified view}
\end{align}
with $\bm{\phi}$ initialized by $\bm{\phi}^{(0)}$, where $\bm{z} \in \{0, 1\}^{|\bm{\phi}^{(0)}|}$. For sparse fine-tuning, the 1s in $\bm{z}$ only locates for the trainable parameters in $\bm{\theta}^{(0)}$, whereas all locations for $\bm{\delta}^{(0)}$ in $\bm{z}$ are 1s for adapter fine-tuning. In a word, the subset of the trainable parameters in $\bm{\phi}$ is fixed for adapter fine-tuning, but task-specific for sparse fine-tuning. 

\textbf{Key Challenges}. Sparse fine-tuning normally gains less attention than adapter fine-tuning. The reasons are two-fold: (1) In the first phase of sparse fine-tuning, the generation of the sparse mask is task-specific, which means different downstream tasks or the same task with different domain data might have different masks, whereas adapter fine-tuning always has fixed positions for adapters; (2) One needs some tricks to generate these masks, like a differential version of $L_0$ norm \cite{DBLP:conf/acl/GuoRK20} or Fisher Information \cite{DBLP:conf/nips/SungNR21}, which requires more computation than a normal full fine-tuning for the same iterations. 

Compared to sparse fine-tuning, adapter fine-tuning typically introduces additional inference latency from the newly added parameters. Though LoRA doesn't have this issue, one can't apply a nonlinear function in the adapter for LoRA since 
\begin{align}
   & \bm{h}\bm{W} + f(\bm{h}\bm{W}_{down})\bm{W}_{up}  \nonumber \\
   \neq & \bm{h}(\bm{W} + f(\bm{W}_{down})\bm{W}_{up})
\label{eq: lora inequality}
\end{align}
, which limits the learning capacity of this method.

\section{Methodologies}
Motivated by the key challenges stated in Section $\S$\ref{sec: preliminaries}, we propose two methods in this section and illustrate them in Figure \ref{fig: pafi hiwi arch}: one for sparse fine-tuning that generates a universal mask for various tasks without any training, and another one for adapter fine-tuning that has the same inference speed as full fine-tuning while requiring even less storage than BitFit \cite{DBLP:conf/acl/ZakenGR22}. 

\subsection{Task-Agnostic Mask Generation}
Compared to adapter fine-tuning with fixed parameters to tune, existing sparse fine-tuning methods typically require extra training to determine which parameters are trainable. This procedure not only requires more computation than full fine-tuning but also hinders the application of this method to federated learning where data is rarely i.i.d. 
Based on this issue, we propose a research question: could we universally select a set of trainable parameters for different tasks?

 To solve this question, we look into the benefit of sequential pre-training and fine-tuning. \citet{DBLP:conf/acl/AghajanyanGZ20} stated that PLM learns generic and distributed enough representations of language to facilitate downstream learning of highly compressed task representation. We hypothesize that the important (in some sense) parameters of a PLM learn a more generic representation of language and therefore favor downstream tasks more than the others. Therefore, we should fix these parameters and only update the unimportant ones. One might argue that the important parameters of a PLM could also be the ones important to downstream tasks and we should train them rather than fix them. We empirically justify our claim in Section $\S$\ref{sec: discussion on pafi}.

Now the question goes to how to select the unimportant parameters so that we can fine-tune them on downstream tasks later. Here we offer two options: one with training on pre-training data and one in a data-less way. Inspired by FISH Mask \cite{DBLP:conf/nips/SungNR21} where the unimportant parameters are the ones with less Fisher information, we can approximate the Fisher information matrix \cite{Fisher1992, DBLP:conf/nips/Amari96} as
\begin{align}
    \hat{\bm{F_{\theta}}} = \frac{1}{N} \sum_{i=1}^N (\triangledown_{\bm{\theta}} \mathcal{L}^{pre}(\mathcal{D}_i; \bm{\theta}))^2
\label{eq: fisher}
\end{align}
where $\hat{\bm{F_{\theta}}} \in \mathbb{R}^{|\bm{\theta}|}$, $\mathcal{D}_i$ is a sample from the pre-training corpus and $\mathcal{L}^{pre}$ is the pre-training objective. An intuitive explanation of Equation \ref{eq: fisher} is: The parameters with a larger value in $\hat{\bm{F_{\theta}}}$ are more important since they cause larger gradient updates. Then the sparse mask comprises the parameters $\{\bm{\theta}_i | \hat{\bm{F}}_{\bm{\theta}_i} \leq sort(\hat{\bm{F}}_{\bm{\theta}})_k \}$, since we only update the unimportant parameters.

Another method is magnitude-based. We simply consider the pre-trained parameters with the smallest absolute magnitude as the unimportant ones, since they contribute the least to the pre-training loss. Then the sparse mask consists of the parameters $\{\bm{\theta}_i | |\bm{\theta}_i| \leq sort(|\bm{\theta}|)_k\}$. 

In this paper, we only explore the magnitude-based method and leave another one for future work. The main reason is: The magnitude-based method doesn't require any training on any data, whereas another method requires the calculation of gradients on pre-training data that are normally large-scale and private for some PLMs. We name our method as PaFi, since it follows a procedure of \textbf{P}runing-\textbf{a}nd-\textbf{Fi}netuning.   

\subsection{Adapter for Pre-trained Parameters}
Compared to sparse fine-tuning that tunes a small portion of existing parameters, adapter fine-tuning introduces new parameters at some fixed positions for different tasks. Normally, the number of added parameters correlates with the complexity of the downstream task. For example, one can add 0.5\% of the PLM parameters to achieve the same performance as full fine-tuning for the GLUE benchmark, while 4\% is required for machine translation \cite{DBLP:conf/iclr/HeZMBN22}. The inference speed of a task is proportional to the number of added parameters, so the introduced inference latency is nonnegligible for complex tasks.

Inspired by LoRA \cite{DBLP:conf/iclr/HuSWALWWC22} which has the same inference speed as full fine-tuning, we propose a new adapter fine-tuning method that applies an adapter directly to pre-trained parameters instead of hidden representations as:
\begin{align}
    \bm{W} \leftarrow \bm{W} + f(\bm{W}\bm{W}_{down})\bm{W}_{up}
\label{eq: hiwi}
\end{align}
If we neglect the nonlinear function, a representation $\bm{h}$ through our linear layer becomes $\bm{hW}(\bm{1} + \bm{W}_{down}\bm{W}_{up})$. Compared to LoRA (See Equation \ref{eq: lora}) where the learned diff matrix is $\triangle \bm{W} = \bm{W}_{down}\bm{W}_{up}$, our learned diff matrix is $\triangle \bm{W} = \bm{W}\bm{W}_{down}\bm{W}_{up}$. Without considering the nonlinear function, the rank of the diff matrix from LoRA is the upper bound of the rank for our diff matrix:
\begin{align}
    & rank(\bm{WW}_{down}\bm{W}_{up})  \nonumber \\
    \leq & min(rank(\bm{W}), rank(\bm{W}_{down}\bm{W}_{up})) \nonumber \\
    = & rank(\bm{W}_{down}\bm{W}_{up})
\end{align}
The equality holds since the rank of a pre-trained weight is empirically larger. To improve the learning capacity (it is related to the matrix rank) of our method, we input a nonlinear function between $\bm{W}_{down}$ and $\bm{W}_{up}$, which is not possible for LoRA if we want to maintain the same inference speed as full fine-tuning (see Equation \ref{eq: lora inequality}).

One obvious advantage of our method is that it has the same inference speed as full fine-tuning, since we can compute $\bm{W} \leftarrow \bm{W} + f(\bm{W}\bm{W}_{down})\bm{W}_{up}$ before the inference step. Another advantage is: we can replace the weight matrix in Equation \ref{eq: hiwi} with the bias term. I.e. we input the pre-trained bias to an adapter to construct a new bias. In this way, we can solve the issue raised by BitFit \cite{DBLP:conf/acl/ZakenGR22}, where the number of bias terms is fixed and therefore BitFit is not scalable. When we apply the adapter to the weight matrix, we need to save $\bm{W}_{down}$ and $\bm{W}_{up}$, since their size are much smaller than $\bm{W}$. However, when we apply the adapter to bias terms, we only need to save the new bias ($\bm{b} \leftarrow \bm{b} + f(\bm{b}\bm{W}_{down})\bm{W}_{up}$) that requires much less storage than saving $\bm{W}_{down}$ and $\bm{W}_{up}$. It also means we require the same storage (for the bias terms) whatever the size of $\bm{W}_{down}$ and $\bm{W}_{up}$, and therefore we can use a large number of trainable parameters. We name our method as HiWi, since it \textbf{Hi}des (throws away) the \textbf{W}e\textbf{i}ghts from adapters.

\section{Experimental Setup}
\subsection{Evaluation Tasks}
Due to limited computation resources, we select six tasks from the GLUE \cite{DBLP:conf/iclr/WangSMHLB19} and SuperGLUE \cite{DBLP:conf/nips/WangPNSMHLB19} benchmarks: two natural language inference tasks (MNLI and RTE), a similarity task (STS-B), a word sense disambiguation task (WiC), a coreference resolution task (WSC) and a causal reasoning task (COPA). For most tasks, we follow the RoBERTa paper \cite{DBLP:journals/corr/abs-1907-11692}, treating MNLI and RTE as sentence-level classification tasks, WiC as a word-level classification task, STS-B as a regression task and WSC as a ranking task. Nevertheless, we implement COPA as a ranking classification task rather than a binary classification task in \citet{DBLP:journals/corr/abs-1907-11692}, since it offers better performance for all methods. 


We term our selected tasks \textit{VariousGLUE}, since they cover a wide range of tasks (classification, ranking, regression) and include high-resource (MNLI), middle-resource (STS-B, WiC and RTE) and low-resource (WSC, COPA) tasks.
For evaluation on VariousGLUE, we report accuracy for MNLI, WiC, RTE, WSC and COPA, and the Pearson correlation coefficient for STS-B on the development sets. More data statistics, implementation details and task selection criteria of VariousGLUE are in Appendix $\S$\ref{sec: Experimental Detail}. 
Except for natural language understanding (NLU) tasks, we also evaluate our methods on a sequence-to-sequence task, i.e. English to Romanian translation with the WMT 2016 En-Ro dataset \cite{DBLP:conf/wmt/BojarCFGHHJKLMN16}, and report BLEU \cite{DBLP:conf/acl/PapineniRWZ02} on the test set.


\begin{table*}[t]
\scriptsize
\centering
\begin{tabular}{l|c|llllllll|l}
    \toprule
    \textbf{Method} & \textbf{\#Tuned} & \textbf{MNLI} & \textbf{QQP} & \textbf{QNLI} & \textbf{SST-2} & \textbf{CoLA} & \textbf{STS-B} & \textbf{MRPC} & \textbf{RTE} & \textbf{Avg}\\
    \hline
    Full FT \cite{DBLP:journals/corr/abs-1907-11692} & 100\% & 90.2 & 92.2 & 94.7 & 96.4 & 68.0 & 92.4 & 90.9 & 86.6 & 88.9 \\
    Our Full FT & 100\% & 90.1\textsubscript{0.09} & \textbf{92.3}\textsubscript{0.00} & \textbf{94.8}\textsubscript{0.05} & 96.4\textsubscript{0.21} & \underline{69.0}\textsubscript{0.82} & 91.9\textsubscript{0.17} & \textbf{91.7}\textsubscript{0.14} & \underline{88.1}\textsubscript{0.63} & \textbf{89.3}\textsubscript{0.26} \\
    \hline
    Linear FT & 0\% & 52.4\textsubscript{0.47} & 75.6\textsubscript{0.19} & 67.4\textsubscript{0.05} & 83.7\textsubscript{0.25} & 00.0\textsubscript{0.00} & 31.2\textsubscript{3.50} & 69.9\textsubscript{0.19} & 54.5\textsubscript{0.14} & 55.1\textsubscript{0.60} \\
    Linear FT\textsubscript{norm} & 0.03\% & 88.4\textsubscript{0.00} & 87.8\textsubscript{0.05} & 92.5\textsubscript{0.12} & 95.1\textsubscript{0.05} & 47.0\textsubscript{0.79} & 75.3\textsubscript{1.89} & 71.1\textsubscript{0.14} & 53.4\textsubscript{0.79} & 76.3\textsubscript{0.48} \\
    Adapter\textsuperscript{\dagger} & 0.2\% & 90.3\textsubscript{0.3} & 91.5\textsubscript{0.1} & \underline{94.7}\textsubscript{0.2} & 96.3\textsubscript{0.5} & 66.3\textsubscript{2.0} & 91.5\textsubscript{0.5} & 87.7\textsubscript{1.7} & 72.9\textsubscript{2.9} & 86.4\textsubscript{1.0} \\
    Adapter\textsuperscript{\dagger} & 1.7\% & 89.9\textsubscript{0.5} & \underline{92.1}\textsubscript{0.1} & \underline{94.7}\textsubscript{0.2} & 96.2\textsubscript{0.3} & 66.5\textsubscript{4.4} & 91.0\textsubscript{1.7} & 88.7\textsubscript{2.9} & 83.4\textsubscript{1.1} & 87.8\textsubscript{1.4} \\
    Pfeiffer Adapter\textsuperscript{\dagger} & 0.2\% & 90.5\textsubscript{0.3} & 91.7\textsubscript{0.2} & \textbf{94.8}\textsubscript{0.3} & \underline{96.6}\textsubscript{0.2} & 67.8\textsubscript{2.5} & 91.9\textsubscript{0.4} & 89.7\textsubscript{1.2} & 80.1\textsubscript{2.9} & 87.9\textsubscript{1.0} \\
    Pfeiffer Adapter\textsuperscript{\dagger} & 0.8\% & 90.2\textsubscript{0.3} & 91.9\textsubscript{0.1} & \textbf{94.8}\textsubscript{0.2} & 96.1\textsubscript{0.3} & 68.3\textsubscript{1.0} & 92.1\textsubscript{0.7} & 90.2\textsubscript{0.7} & 83.8\textsubscript{2.9} & 88.4\textsubscript{0.8} \\
    LoRA \cite{DBLP:conf/iclr/HuSWALWWC22} & 0.2\% & \textbf{90.6}\textsubscript{0.2} & 91.6\textsubscript{0.2} & \textbf{94.8}\textsubscript{0.3} & 96.2\textsubscript{0.5} & 68.2\textsubscript{1.9} & \underline{92.3}\textsubscript{0.5} & 90.2\textsubscript{1.0} & 85.2\textsubscript{1.1} & \underline{88.6}\textsubscript{0.7} \\
    Diff Pruning & 0.5\% & \underline{90.3}\textsubscript{0.08} & 90.3\textsubscript{0.17} & 94.6\textsubscript{0.22} & 96.4\textsubscript{0.21} & 65.1\textsubscript{2.39} & 92.0\textsubscript{0.17} & 90.2\textsubscript{1.11} & 84.5\textsubscript{1.18} & 87.9\textsubscript{0.69}\\
    FISH Mask & 0.5\% & 90.2\textsubscript{0.08} & 89.8\textsubscript{0.17} & 94.1\textsubscript{0.26} & 96.1\textsubscript{0.38} & 66.3\textsubscript{1.89} & \textbf{92.5}\textsubscript{0.08} & 88.7\textsubscript{0.92} & 86.3\textsubscript{1.07} & 88.0\textsubscript{0.61} \\
    \hline
    PaFi & 0.5\% & 90.2\textsubscript{0.05} & 90.3\textsubscript{0.05} & 94.6\textsubscript{0.08} & \textbf{96.7}\textsubscript{0.19} & \textbf{70.2}\textsubscript{0.46} & 91.9\textsubscript{0.24} & \underline{91.4}\textsubscript{0.14} & \textbf{88.8}\textsubscript{0.63} & \textbf{89.3}\textsubscript{0.23}\\
    \bottomrule
\end{tabular}
\caption{\label{tab: sparse fine-tuning}
Sparse fine-tuning on GLUE. The best and second-best scores are in \textbf{bold} and \underline{underlined}, respectively. Results of methods with ``\dagger'' are also copied from \citet{DBLP:conf/iclr/HuSWALWWC22}. The tasks are ordered with their number of samples from largest to smallest.}
\end{table*}

\subsection{Baselines}
\label{sec: baselines}
To compare with other baselines broadly, we replicate their setups since most of them are not evaluated on the same tasks or use the same PLM (RoBERTa\textsubscript{LARGE}) as ours. If possible, we also report their scores.  

\textbf{Full fine-tuning (Full FT)} updates all parameters. \textbf{Linear fine-tuning (Linear FT)} only tunes the added classifier. \textbf{Linear fine-tuning with normalization (Linear FT\textsubscript{norm}}) fine-tunes the classifier and all normalization layers of the PLM. We borrow the fine-tuning recipe from \citet{DBLP:journals/corr/abs-1907-11692} for these three baselines. 
  
Both \textbf{Diff Pruning} \cite{DBLP:conf/acl/GuoRK20} and \textbf{FISH Mask} \cite{DBLP:conf/nips/SungNR21} are chosen as sparse fine-tuning baselines. We implement them on their own frameworks\footnote{\url{https://github.com/dguo98/DiffPruning}; \url{https://github.com/varunnair18/FISH}} with their own recipes (combined with our recipe for middle-/low-resource tasks). 

We select three adapter variants as our baselines: \textbf{Adapter} \cite{DBLP:conf/icml/HoulsbyGJMLGAG19}, \textbf{Pfeiffer Adapter} \cite{DBLP:conf/eacl/PfeifferKRCG21} and \textbf{LoRA} \cite{DBLP:conf/iclr/HuSWALWWC22}. In addition, we also compare our methods to \textbf{BitFit} \cite{DBLP:conf/acl/ZakenGR22}, \textbf{Prefix Tuning} \cite{DBLP:conf/acl/LiL20} and \textbf{MAM Adapter} \cite{DBLP:conf/iclr/HeZMBN22}. MAM Adapter combines prefix tuning and adapter, offering a new state-of-the-art.

If not specified otherwise, we reproduce these baselines on RoBERTa\textsubscript{LARGE} with our own training recipe (see Section $\S$\ref{sec: implementation}), if they don't offer results on our selected tasks or use different PLMs. You can find more details about the calculation of trainable parameters and storage requirements of these methods in Appendix $\S$\ref{sec: number of tunable parameters and storage}.

\subsection{Implementation}
\label{sec: implementation}
We use the encoder-only RoBERTa\textsubscript{LARGE} model \cite{DBLP:journals/corr/abs-1907-11692} as the underlying model for all NLU tasks and the encoder-decoder mBART\textsubscript{LARGE} model \cite{DBLP:journals/tacl/LiuGGLEGLZ20} for MT. All our implementations are on the Fairseq framework \cite{DBLP:conf/naacl/OttEBFGNGA19}. For NLU tasks, we sweep learning rates in \{3, 4, 5, 6, 7\} $\cdot 10^{-4}$ (inspired by the best results obtained in LoRA \cite{DBLP:conf/iclr/HuSWALWWC22}), batch sizes in \{16, 32\}, and the number of epochs in \{10, 20\} (for tasks with the number of samples over 100K, we only train for 10 epochs). Other settings of the optimizer stay the same as the RoBERTa paper. For the En-Ro task, we borrow the same training recipe from \citet{DBLP:conf/iclr/HeZMBN22}, i.e. setting the learning rate as $5\cdot10^{-5}$, a mini-batch with 16384 tokens, a label smoothing factor of 0.1 \cite{DBLP:conf/cvpr/SzegedyVISW16, DBLP:conf/coling/GaoLN20} for 50K iterations. 

We run all experiments on a single NVIDIA RTX A6000 GPU with 48G memory. 
In addition, we run the same task of a method in the above-mentioned grid search space three times with different random seeds, choose the best result from each run, and report the median and standard deviation of these three best results. 

\textbf{PaFi and HiWi.} By default, we select the bottom-k parameters for PaFi group-wise rather than globally. I.e. we select $k$ parameters with the smallest absolute magnitude within each group (a weight matrix or a bias term is considered as a group) and only fine-tune them. In addition, we fine-tune all parameters from normalization layers and don't update the token and position embeddings at all. More discussion about this setting is in Section $\S$\ref{sec: discussion on pafi}. For HiWi, the default setting is feeding the bias rather than the weight to an adapter, because it requires much less storage.

\begin{table*}[t]
\centering
\small
\begin{tabular}{l|c c|llllll|l}
    \toprule
    \textbf{Method} & \textbf{\#Tuned} & \textbf{\#Stored} & \textbf{MNLI} & \textbf{WiC} & \textbf{STS-B} & \textbf{RTE} & \textbf{WSC} & \textbf{COPA} & \textbf{Avg}\\
    \hline
    Full FT\textsuperscript{\dagger} & 100\% & 100\% & 90.2 & 75.6 & 92.4 & 86.6 & - & 94.0 &  - \\
    Our Full FT & 100\% & 100\% & 90.1\textsubscript{0.09} & \textbf{74.0}\textsubscript{0.41} & \underline{91.9}\textsubscript{0.17} & 88.1\textsubscript{0.63} & 87.5\textsubscript{1.09} & \textbf{96.0}\textsubscript{0.47} & \underline{88.0}\textsubscript{0.48} \\
    \hline
    Linear FT & 0\% & 0\% & 52.4\textsubscript{0.47} & 67.6\textsubscript{0.46} & 31.2\textsubscript{3.50} & 54.5\textsubscript{0.14} & 68.3\textsubscript{0.00} & 72.0\textsubscript{1.63} & 58.7\textsubscript{1.04} \\
    Linear FT\textsubscript{norm} & 0.03\% & \textbf{0.03}\% & 88.4\textsubscript{1.42} & 67.9\textsubscript{0.33} & 75.3\textsubscript{1.89} &  53.4\textsubscript{0.79} & 75.9\textsubscript{0.85} & 74.0\textsubscript{1.41} & 73.0\textsubscript{0.88} \\
    BitFit & 0.08\% & 0.08\% & 89.5\textsubscript{0.05} & 71.9\textsubscript{0.45} & 91.4\textsubscript{0.09} & 88.1\textsubscript{0.33} & 85.7\textsubscript{1.53} & 89.0\textsubscript{1.24} & 85.9\textsubscript{0.62}\\
    Prefix Tuning & 0.5\% & 0.5\% & 89.9\textsubscript{0.12} & 69.7\textsubscript{0.62} & 91.4\textsubscript{0.75} & 76.5\textsubscript{2.1} & \textbf{91.1}\textsubscript{1.59} & 74.0\textsubscript{1.89} & 82.1\textsubscript{1.18} \\
    Adapter & 0.5\%  & 0.5\% & \underline{90.6}\textsubscript{0.16} & 71.9\textsubscript{0.12} & \textbf{92.1}\textsubscript{0.21} & 87.4\textsubscript{0.45} & 84.8\textsubscript{1.00} & 88.0\textsubscript{2.16} & 85.8\textsubscript{0.69} \\
    Pfeiffer Adapter & 0.5\% & 0.5\% & \textbf{90.8}\textsubscript{0.08} & 71.9\textsubscript{0.47} & \textbf{92.1}\textsubscript{0.26} & 88.4\textsubscript{0.46} & 86.7\textsubscript{0.40} & 90.0\textsubscript{0.94} & 86.6\textsubscript{0.44} \\
    LoRA & 0.5\% & 0.5\% & \underline{90.6}\textsubscript{0.17} & 72.3\textsubscript{0.68} & 91.7\textsubscript{0.24} & 88.4\textsubscript{1.02} & 88.4\textsubscript{0.42} & 92.0\textsubscript{2.36} & 87.2\textsubscript{0.81} \\
    MAM Adapter & 0.5\% & 0.5\% & \underline{90.6}\textsubscript{0.17} & 72.4\textsubscript{0.79} & \textbf{92.1}\textsubscript{0.05} & \textbf{89.5}\textsubscript{1.09} & 88.3\textsubscript{0.05} & 92.0\textsubscript{0.47} & 87.5\textsubscript{0.44} \\
    \hline
    PaFi & 0.5\% & 0.5\% & 90.2\textsubscript{0.05} & 72.6\textsubscript{0.14} & \underline{91.9}\textsubscript{0.24} & \underline{88.8}\textsubscript{0.63} & 85.2\textsubscript{0.45} & 92.0\textsubscript{1.70} & 86.8\textsubscript{0.54} \\
    HiWi (r=4) & 0.5\% & \textbf{0.03}\% & 90.2\textsubscript{0.05} & 73.4\textsubscript{0.83} & 91.6\textsubscript{0.05} & 88.1\textsubscript{0.05} & 87.5\textsubscript{0.42} & 93.0\textsubscript{0.00} & 87.3\textsubscript{0.23} \\
    HiWi (r=16) & 2.0\% & \textbf{0.03}\% & 90.2\textsubscript{0.09} & \underline{73.7}\textsubscript{0.33} & \underline{91.9}\textsubscript{0.17} & \underline{88.8}\textsubscript{0.33} & \underline{89.3}\textsubscript{2.25} & \underline{95.0}\textsubscript{1.25} & \textbf{88.2}\textsubscript{0.74}\\  
    \bottomrule
\end{tabular}
\caption{\label{tab: all}
Results on VariousGLUE. The tasks are ordered with their number of samples from largest to smallest.  The best and second-best scores are in \textbf{bold} and \underline{underlined}, respectively. Results of the method with ``\dagger'' are copied from \citet{DBLP:journals/corr/abs-1907-11692}. Notably, the storage of HiWi is invariant to the number of trainable parameters.}
\end{table*}

\begin{table}
\centering
\small
\begin{tabular}{lccl}
    \toprule
    \textbf{Method} & \textbf{\#Tuned} & \textbf{\#Stored} &  \textbf{BLEU} \\
    \hline
    Full FT\textsuperscript{\dagger} & 100\% & 100\% & 37.3 \\
    \hline
    BitFit\textsuperscript{\dagger} & 0.05\% & 0.05\% & 26.4 \\
    Prefix Tuning\textsuperscript{\dagger} & 10.2\% & 2.4\% & 35.6 \\
    Pfeiffer Adapter\textsuperscript{\dagger} & 4.8\% & 4.8\% & 36.9\textsubscript{.1} \\
    LoRA(ffn)\textsuperscript{\dagger} & 4.1\% & 4.1\%  & 36.8\textsubscript{.3} \\
    MAM Adapter\textsuperscript{\dagger} & 14.2\% & 4.5\% & 37.5\textsubscript{.1} \\
    \hline
    PaFi & 4.5\% & 4.5\% & \underline{37.7}\textsubscript{.1} \\
    PaFi & 14.2\% & 14.2\% & \textbf{38.3}\textsubscript{.1} \\
    HiWi for Bias & 4.7\%  & 0.02\% & 28.0\textsubscript{.2} \\
    HiWi for Weight & 4.7\% & 4.7\% & 36.9\textsubscript{.2} \\
    \bottomrule
\end{tabular}
\caption{\label{tab: translation}
Results on WMT 2016 En-Ro. The best and second-best scores are in \textbf{bold} and \underline{underlined}, respectively. Results of the methods with ``\dagger'' are copied from \citet{DBLP:conf/iclr/HeZMBN22}. $r=64$ for HiWi.}
\end{table}

\section{Result and Discussion}
In this section, we present the results of baselines and our proposed methods on GLUE, VariousGLUE and translation tasks. 

\subsection{Sparse Fine-Tuning on GLUE} Since the frameworks of Diff Pruning and FISH Mask only support the GLUE benchmark, we evaluate our proposed sparse fine-tuning method, PaFi, on GLUE and show the results in Table \ref{tab: sparse fine-tuning}, where we follow the sparsity setting of Diff Pruning and FISH Mask, and set it as 0.5\%. In the RoBERTa paper \cite{DBLP:journals/corr/abs-1907-11692}, some low-resource tasks (RTE, MRPC and STS-B) are initialized from the fine-tuned model on MNLI rather than the PLM. We don't follow this setup and always consider the PLM as an initialization as \citet{DBLP:conf/icml/HoulsbyGJMLGAG19}. Surprisingly, our reproduction outperforms the reported score in the RoBERTa paper by 0.4\%.

Compared to existing sparse and adapter fine-tuning methods, our PaFi obtains the best result, achieving the same score (89.3) as Full FT with only updating 0.5\% parameters. PaFi also outperforms Diff Pruning and FISH Mask by a significant margin, with at least a 1.3\% improvement on average, which demonstrates the effectiveness of our simple parameter selection method. 

In addition, PaFi also shows its efficiency in terms of storage and training compared to other sparse fine-tuning methods. The saving of the indices for updated parameters consumes the same memory as the saving of these parameters if both are in fp32. One only needs to save one mask for PaFi, but the same number of masks as the number of tasks for Diff Pruning and FISH Mask. Our one-mask-for-all setting is also suitable for federated learning, where non-i.i.d. data could use the same mask, which is impossible for Diff Pruning and FISH Mask. For training costs, PaFi doesn't require any training in the mask generation phase, while Diff Pruning and FISH Mask require more computation ($>2$ times for Diff Pruning) than Full FT because of the calculation of differential $L_0$ norm or Fisher information matrix. Specifically on the MNLI task, Diff Pruning, FISH Mask and PaFi spend 19.9h, 1m39s and 2s to generate the sparse mask, respectively. And both Diff Pruning and FISH Mask require the data of downstream task for the mask generation, while the mask generation of PaFi is data-less.

\subsection{Results on VariousGLUE}
Table \ref{tab: all} shows the results of different methods on VariousGLUE. For most methods, the number of trainable parameters stays the same as the number of stored parameters. However, Prefix Tuning and MAM Adapter apply a re-parameterization trick and throw away some trainable parameters after training, which might result in different numbers (see Appendix $\S$\ref{sec: number of tunable parameters and storage}). In addition, the storage requirement of HiWi is invariant to the number of trainable parameters since we throw away all adapter parameters and only save the new bias terms.

Compared to Table \ref{tab: sparse fine-tuning}, PaFi performs unexpectedly worse, which justifies the necessity of evaluation on various tasks for PEFT methods. On high-resource (MNLI) and most middle-resource (STS-B and RTE) tasks, PaFi is on par with Full FT. Though not the best, PaFi still outperforms some adapter fine-tuning methods (Adapter and Pfeiffer Adapter), which shows tuning existing parameters is enough for some tasks.

When using the same number of trainable parameters (0.5\%), HiWi performs on par with the best baseline, MAM Adapter (87.3 vs 87.5). Notably, MAM Adapter is an additive work that combines Prefix Tuning and adapters together. We could also implement HiWi and Prefix Tuning in the same framework and leave this exploration to future work. If we increase the number of trainable parameters to 2\% (almost requiring the same training time and memory footprint as 0.5\%), HiWi outperforms all baselines and Full FT (88.2 vs 88.0). Notably, HiWi only requires a fixed storage, around 0.03\% of the PLM total parameters, which is also the least storage requirement among all methods.


\begin{figure*}[ht]
  \centering
    \includegraphics[width=0.98\textwidth]{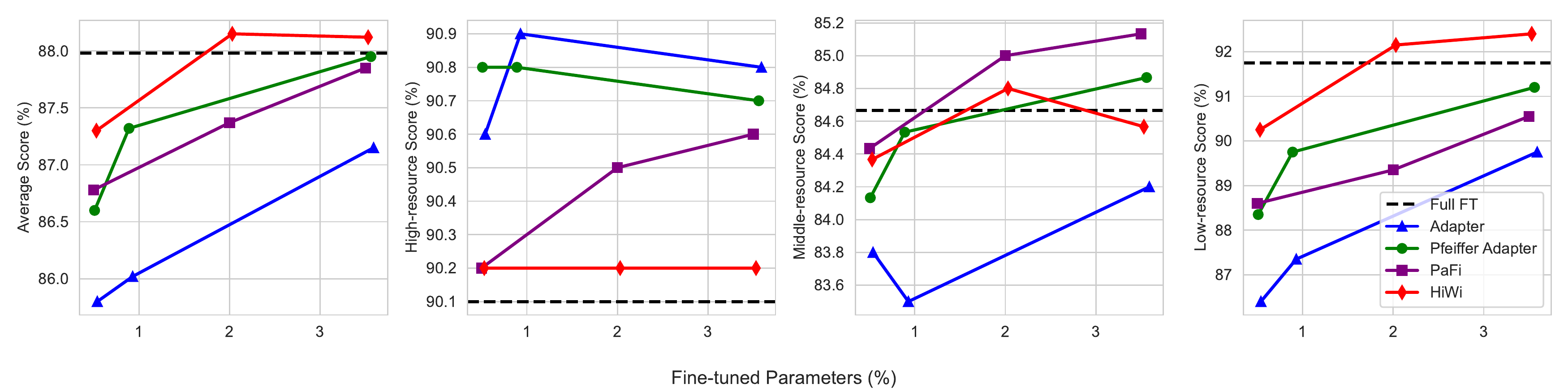}
    \caption{Scalability of different methods: (a) Average performance on VariousGLUE. (b)(c)(d): average performance on high-resource (MNLI), middle-resource (WiC, STS-B, RTE) and low-resource (WSC, COPA) tasks.}
    \label{fig: scale}
\end{figure*}

\label{sec: ablation}
\begin{table}
\small
\centering
\begin{tabular}{l|lll}
    \toprule
    \textbf{Method} & \textbf{WiC} & \textbf{RTE} & \textbf{COPA} \\
    \hline
    Full FT & 74.0\textsubscript{0.41} & 88.1\textsubscript{0.63} & 96.0\textsubscript{0.47} \\
    \hline
    \cellcolor{Gray}Smallest &  72.6\textsubscript{0.14} & \textbf{88.8}\textsubscript{0.63} & \textbf{92.0}\textsubscript{1.70} \\
    Largest & \textbf{72.9}\textsubscript{0.54} & 84.1\textsubscript{0.66} & 90.0\textsubscript{0.81} \\
    Middle & 72.7\textsubscript{0.45} & 83.8\textsubscript{0.85} & 90.0\textsubscript{0.82} \\
    Random & 71.9\textsubscript{0.40} & 84.5\textsubscript{1.80} & 91.0\textsubscript{0.94} \\
    \hline
    Not tune norm & 72.4\textsubscript{0.57} & 88.1\textsubscript{0.45} & 90.0\textsubscript{0.94} \\
    Tune embed & 72.6\textsubscript{0.24} & 88.4\textsubscript{0.14} & 91.0\textsubscript{0.94} \\
    \bottomrule
\end{tabular}
\caption{\label{tab: sparse ablation}
Ablation studies of PaFi on: (1) How to choose trainable parameters; (2) Which groups should be tuned. The method in \textcolor{gray}{gray} is PaFi with the default setting.}
\end{table}

\subsection{Results on MT}
Compared to NLU tasks, where the number of trainable parameters is negligible, MT task requires more trainable parameters to achieve a similar result as Full FT. We show the results of different methods on WMT 2016 En-Ro in Table \ref{tab: translation}. PaFi performs the best and outperforms Full FT (37.7 vs 37.3 with 4.5\% parameters). Similar to the conclusion that we draw from Table \ref{tab: all}, PaFi is good at high- and middle-resource tasks. 

HiWi for bias works unexpectedly worse, while HiWi for weight is on par with LoRA. We argue: Though we can improve the learning capacity of bias terms with adapters, the limited amount of biases still hinders HiWi's representative ability. HiWi for weight solves this issue by feeding a much larger amount of weights to adapters. In addition, most baselines, except for BitFit and LoRA, introduce nonnegligible inference latency (proportional to the stored parameters), while PaFi and HiWi share the same inference speed as Full FT. Specifically, the inference time on En-Ro is 110s for Full FT, LoRA, HiWi and PaFi, while it's 125s for Pfeiffer Adapter (13.6\% more inference latency).

\subsection{Scalability}
Not all PEFT methods benefit monotonically from having more trainable parameters. \citet{DBLP:conf/acl/LiL20} and \citet{DBLP:conf/iclr/HuSWALWWC22} have shown that Prefix Tuning can't be scaled up well. Here we investigate the scalability of our methods and show the results in Figure \ref{fig: scale}. On average, all listed methods could be scaled well, and HiWi always outperforms other methods. However, these methods show different scaling behaviors for tasks in different resources. 

For the high-resource task, HiWi performs the worst and is stable with an increasing number of trainable parameters, while PaFi shows a well-behaved pattern. This also explains the best performance of PaFi on En-Ro. I.e. PaFi could obtain stronger performance for the setting of a high-resource task and a high number of trainable parameters. For middle-resource tasks, PaFi outperforms other methods and still has a good scaling behavior. For low-resource tasks, HiWi performs the best. The best overall performance of HiWi mainly benefits from these low-resource tasks, which shows HiWi is an effective option for the low-resource or few-shot learning setting.

In summary, PaFi shows its superiority when the task is high-resource and the number of tunable parameters is not too small, while the superiority of HiWi locates in the low-resource (for tasks and the number of tunable parameters) setting. 

\subsection{The Default Setting of PaFi}
\label{sec: discussion on pafi}
Table \ref{tab: sparse ablation} shows our investigation of the PaFi's settings. We implement four ablation experiments on how to choose the trainable parameters. Overall, updating the parameters with the smallest absolute magnitudes offers the best results. The gap between our default setting and other options is the largest for RTE. In addition, tuning the normalization layers is necessary for all tasks. 

The results of tuning embeddings are similar to the results of without tuning embeddings, but a little lower. According to Figure \ref{fig: weight dist} (in Appendix), the embedding layer always has a higher mean, twice as the mean from other layers. This is also one reason why we don't tune parameters from the embedding layer, since most of them have a higher absolute magnitude, showing that they are important. Another reason is that the embedding layer occupies a large number of parameters. We can spare the budget for trainable parameters from this layer and share it with other layers. 


\section{Related Works}
\label{sec: detailed related work}
\textbf{Parameter-Efficient Fine-Tuning for PLMs}. Sparse fine-tuning \cite{DBLP:conf/acl/ZakenGR22, DBLP:conf/acl/GuoRK20, DBLP:conf/nips/SungNR21} offers a tool to explore the over-parameterization of PLMs and doesn't introduce additional latency. Infused fine-tuning \cite{DBLP:conf/icml/HoulsbyGJMLGAG19, DBLP:conf/eacl/PfeifferKRCG21, DBLP:conf/iclr/HuSWALWWC22, DBLP:conf/iclr/HeZMBN22} is highly modularized and has a lower task-switching overhead during inference. Our PaFi simplifies existing sparse fine-tuning methods and makes it more practical for the communication-frequent setting (federated learning). Our HiWi lowers the requirement for storage to a scale and doesn't induce any inference latency. 

Other efficient parameterization methods, like COMPACTER \cite{DBLP:conf/nips/MahabadiHR21}, efficiently parametrize the adapter layers and reduce the number of trainable parameters. They are orthogonal to our work and can be combined with our HiWi. A recently proposed activation method, (IA)$^3$ \cite{DBLP:journals/corr/abs-2205-05638}, achieves a new state-of-the-art on few-short learning and has a similar storage requirement as HiWi. However, it can't be scaled up and performs worse than adapter fine-tuning methods when the number of samples is not too small\footnote{\url{https://adapterhub.ml/blog/2022/09/updates-in-adapter-transformers-v3-1/}}.  

\textbf{Pruning.} Network pruning is a technique for sparsifying neural networks while sacrificing minimal performance \cite{DBLP:journals/corr/abs-1801-05787, DBLP:conf/icml/LiuZKZXWCYLZ21}. Though we borrow some methods from it, like parameter selection with magnitude or Fisher information, we don't prune any parameter and maintain a similar performance as full fine-tuning. 

\section{Conclusion}
In this work, we first propose PaFi as a novel but simple method for computing the sparse mask without any training and in a task-agnostic way. It selects the parameters with the lowest absolute magnitude from PLMs and tunes them, with keeping others frozen. We demonstrate the effectiveness of PaFi on the GLUE benchmark and translation task. Secondly, we propose HiWi as an adapter fine-tuning method that feeds pre-trained parameters instead of hidden representations to the adapters. It doesn't introduce any inference latency. Furthermore, it requires the lowest storage while outperforming other strong baselines. In the future, we will try to compute the sparse mask with Fisher information and estimate our methods in a realistic few-shot learning setting. 

\section*{Acknowledgements}

This research was partly supported by the Netherlands Organization for Scientific Research (NWO) under project number VI.C.192.080. We also gratefully acknowledge the valuable feedback from ACL2023 reviewers.

\section*{Limitations}
We acknowledge the main limitation of this work is that we only evaluate our methods on some tasks from the GLUE and SuperGLUE benchmarks due to limited computation resources. And all tasks are not in a realistic few-shot setting, where the number of training samples is less than a few hundred and development sets are not offered. The benefit of PEFT methods could come from an exhaustive search of hyper-parameters for the development sets, while the realistic few-shot setting could solve this issue and shed more light on PEFT. It would be interesting to see how our methods and other baselines perform on a wide range of few-shot tasks.

In addition, current frameworks are not friendly for sparse fine-tuning methods. Most works (Diff Pruning, FISH Mask and our PaFi) still need to calculate a full gradient of all parameters and selectively update the masked parameters, which makes it cost the same training time as full fine-tuning. 

Last but not least, we only estimate our methods on one single complex task, i.e. WMT 2016 En-Ro. One might not draw the same conclusion as ours on other complex tasks, like machine translation for different languages and resources, summarization tasks, and so on.

\section*{Ethics Statement}
The PLMs involved in this paper are RoBERTa\textsubscript{LARGE} and mBART\textsubscript{LARGE}. RoBERTa\textsubscript{LARGE} was pre-trained on BookCorpus \cite{DBLP:conf/iccv/ZhuKZSUTF15}, CC-News \cite{ccnews}, OpenWebText \cite{opentext} and Stories \cite{DBLP:journals/corr/abs-1806-02847}. mBART\textsubscript{LARGE} was pre-trained on CC25 extracted from Common Crawl \cite{DBLP:conf/lrec/WenzekLCCGJG20, DBLP:conf/acl/ConneauKGCWGGOZ20}. All our models may inherit biases from these corpora.


\bibliography{anthology,custom}
\bibliographystyle{acl_natbib}

\newpage
\appendix

\begin{figure}[h]
  \centering
    \includegraphics[width=0.48\textwidth]{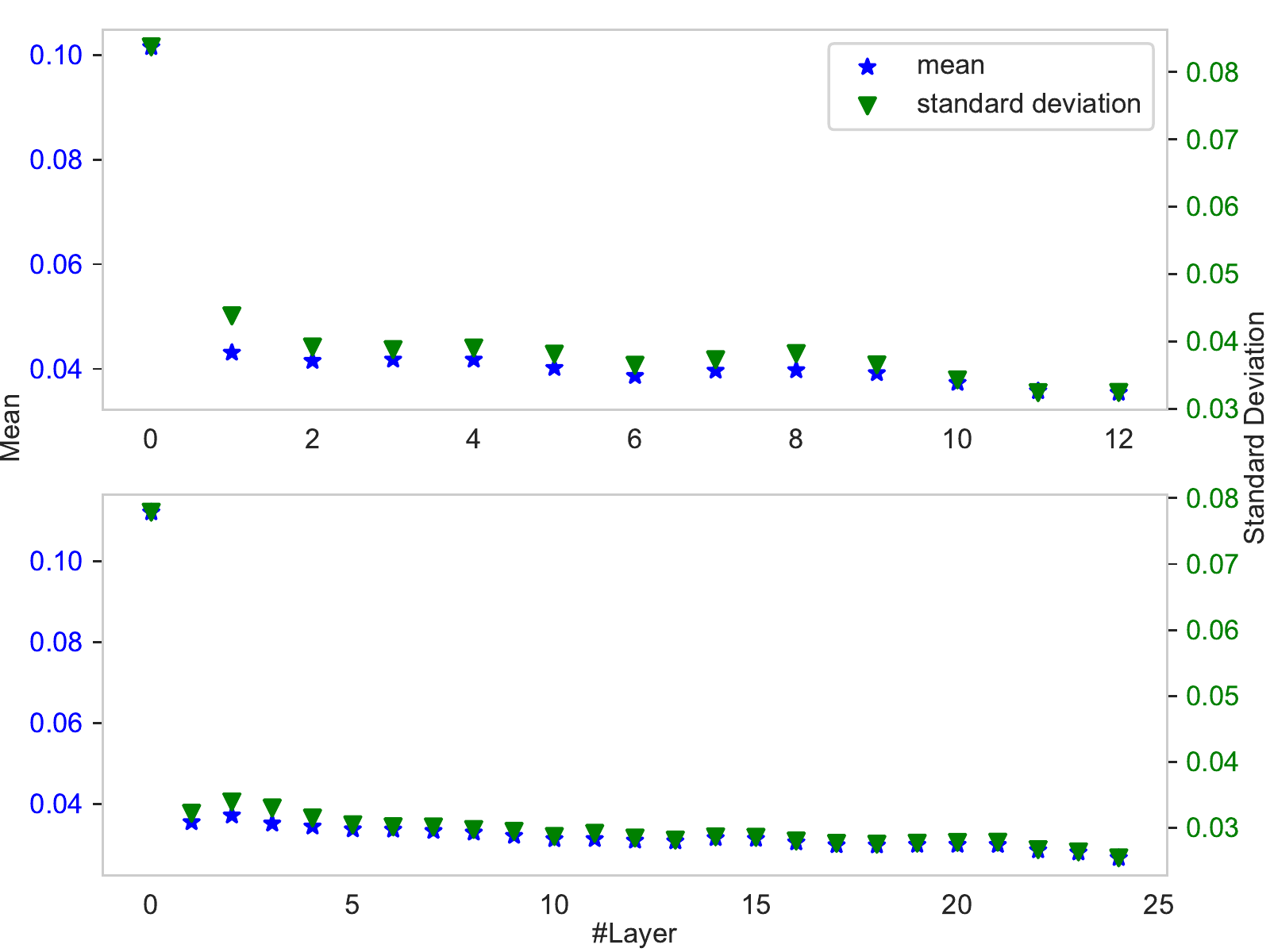}
    \caption{Mean and standard deviation of parameters' absolute magnitudes from different layers. \textbf{Up}: RoBERTa\textsubscript{BASE}, \textbf{Bottom}: RoBERTa\textsubscript{LARGE}. The 0\textsuperscript{th} layer denotes the token and position embeddings.}
    \label{fig: weight dist}
\end{figure}

\section{Experimental Detail}
\label{sec: Experimental Detail}

\subsection{Data Statistics}
We show the statistics of VariousGLUE and En-Ro in Table \ref{tab: data statistics}. We test our methods on four types of tasks that are high-resource (MNLI and En-Ro), middle-resource (STS-B, WiC and RTE) or low-resource (WSC and COPA).

\subsection{Implementation Details of VariousGLUE}
Due to limited computation resources, we could not evaluate our methods and baselines on all tasks from the GLUE \cite{DBLP:conf/iclr/WangSMHLB19} and SuperGLUE \cite{DBLP:conf/nips/WangPNSMHLB19} benchmarks. We select six tasks from these two benchmarks. Except for COPA, we follow the same implementation as RoBERTa \cite{DBLP:journals/corr/abs-1907-11692}. We list the details as follows:
\begin{itemize}
    \item \textbf{MNLI} \cite{DBLP:conf/naacl/WilliamsNB18} and \textbf{RTE} \cite{DBLP:conf/mlcw/DaganGM05, rte2, DBLP:conf/acl/GiampiccoloMDD07, DBLP:conf/tac/BentivogliMDDG09}: The input format is ``[CLS] sentence$_1$ [SEP] [SEP] sentence$_2$ [SEP]''. We input the representation of the [CLS] token from the encoder to a multi-class classifier for prediction. 
    \item \textbf{WiC} \cite{DBLP:conf/naacl/PilehvarC19}: WiC has the same input format as MNLI and RTE. We feed the concatenation of the representation of the two marked words and the [CLS] token to a binary classifier. 
    \item \textbf{STS-B} \cite{DBLP:journals/corr/abs-1708-00055}: STS-B also has the same input format as MNLI and RTE. We feed the representation of the [CLS] token to a regression layer (similar to the classification layer with only one class).
    \item \textbf{WSC} \cite{DBLP:conf/kr/LevesqueDM12}: We first detect all noun phrases from the sentence. Supposed $n$ noun phrases are detected, we replace the pronoun with these phrases to construct $n$ new sentences and input them to RoBERTa in a batch way in the format of ``[CLS] sentence [SEP]''. After the masked word prediction layer, we take the corresponding logits for these $n$ noun phrases. Some noun phrases might be a span. We average the logits in this span to obtain a single logit for each noun phrase. For the sample that offers a correct match between the noun phrase and the pronoun, we assign 1 as a label to the logit of this noun phrase and 0 to the other and calculate the cross-entropy loss. Even though we have to throw away the annotated samples that have incorrect matches, this method offers the best result. During inference, the noun phrase with the biggest logit is the prediction. 
    \item \textbf{COPA} \cite{DBLP:conf/aaaiss/RoemmeleBG11} The input format for a single sample is ``[CLS] Because sentence$_1$, so sentence$_2$ [SEP]'' or ``[CLS] Because sentence$_2$, so sentence$_1$ [SEP]''. We feed these two inputs from the same sample to RoBERTa in a batch way, then input the representation of both [CLS]s to a classifier with only one class, making sure the logit from the input with the correct causal effect is larger than another one by calculating the cross-entropy loss.   
\end{itemize}
Except for WSC, we always insert a classifier layer on top of the RoBERTa encoder. For WSC, we use the original masked word prediction layer from the PLM and keep it frozen. You can find more implementation details in our codebase.

\begin{table}[t]
\centering
\begin{tabular}{llll}
    \toprule
    \textbf{Type} & \textbf{Name} & \textbf{\#Train} & \textbf{\#Dev.}\\
    \hline
    \multirow{3}{*}{classification} & MNLI & 392702 & 19647\\
    & WiC & 5428 & 638 \\
    & RTE & 2490 & 277 \\
    \hline
    regression & STS-B & 5749 & 1500 \\
    \hline
    \multirow{2}{*}{ranking} & WSC & 554 & 104 \\
    & COPA & 400 & 100 \\
    \hline
    translation & En-Ro & 610320 & 1999  \\
    \bottomrule
\end{tabular}
\caption{\label{tab: data statistics}
Statistics of evaluation tasks. The test set of En-Ro also has 1999 samples.
}
\end{table}

\begin{figure*}
  \centering
    \includegraphics[width=0.98\textwidth]{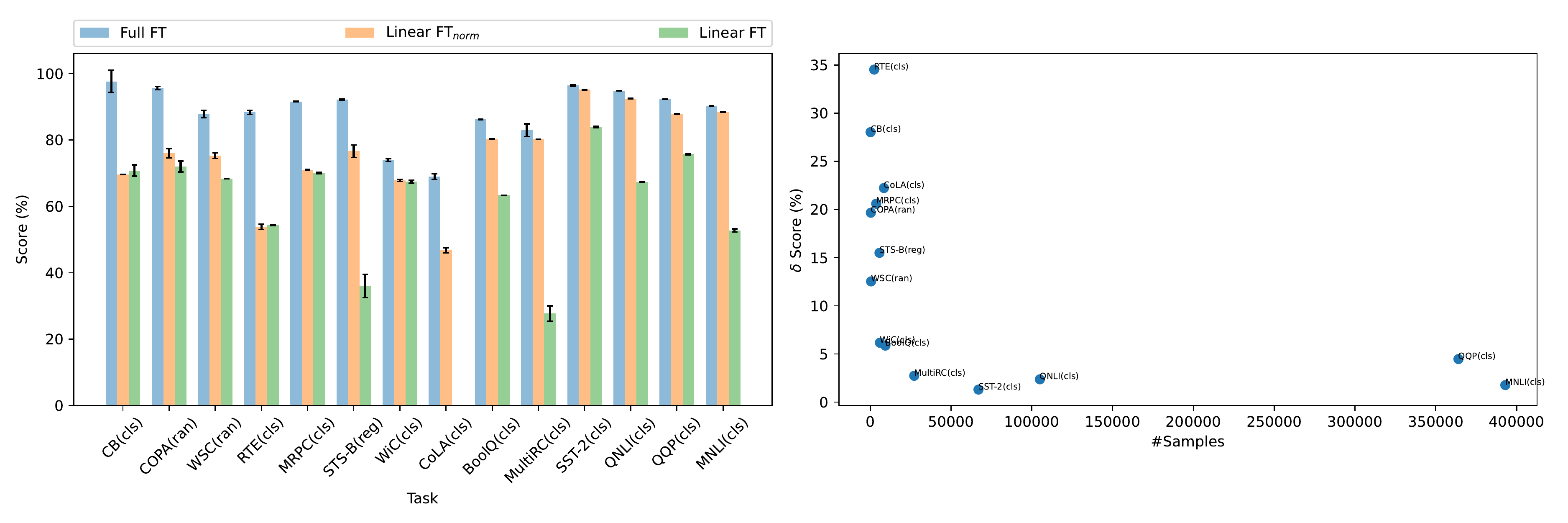}
    \caption{Task Selection. \textbf{Left}: The performance of the tasks from the GLUE and SuperGLUE benchmarks for different methods. The tasks are ordered with their number of samples from smallest to largest. The short terms in the brackets denote the task type. ``cls'', ``ran'' and ``reg'' are the short terms of classification, ranking and regression, respectively. \textbf{Right}: Performance difference between Linear FT\textsubscript{norm} and Full FT.}
    \label{fig: task selection}
\end{figure*}

\begin{table}[t]
\centering
\begin{tabular}{ccc}
    \toprule
    Full FT & Linear FT\textsubscript{norm} & Linear FT \\
    88.5\textsubscript{0.67} & 75.8\textsubscript{0.47} & 57.8\textsubscript{0.78} \\
    \bottomrule
\end{tabular}
\caption{\label{tab: average over all tasks}
The average performance on all tasks.
}
\end{table}

\subsection{Criteria for Task Selection}
Most PEFT methods are evaluated on the GLUE benchmark. We argue that the GLUE tasks might be too easy for PLMs, since all GLUE tasks, except for STS-B, are classification tasks and some PLMs outperform our humans by a large margin according to the GLUE leaderboard\footnote{\url{https://gluebenchmark.com/leaderboard}}. In addition, the number of training samples for each GLUE task is big (RTE is the task with the minimum number of samples, 2.5K). Most works have to construct few-shot learning tasks from GLUE by themselves. Compared to GLUE, the SuperGLUE benchmark offers more low-resource tasks and is much more difficult for PLMs. Due to computational resource limits, we can't evaluate on all GLUE and SuperGLUE tasks, and therefore want to select some tasks from these two benchmarks.

Our selection criteria are three-dimensional: complexity, variety in task type and variety in task resource. Deciding whether a task is easy or complex could be subjective. Since our main research topic in this paper is PEFT, we determine the complexity of a task with its performance improvement from Linear FT\textsubscript{norm} to Full FT (see Section $\S$\ref{sec: baselines} for these baselines). Linear FT\textsubscript{norm} requires the minimum trainable parameters among all PEFT methods. If a task obtains a huge improvement from Linear FT\textsubscript{norm} to Full FT, it means this task needs many parameters to tune, and therefore it is difficult for PEFT methods and a complex task.

Figure \ref{fig: task selection} shows the performance of Full FT, Linear FT and Linear FT\textsubscript{norm} on the tasks from GLUE and SuperGLUE. Overall, Linear FT\textsubscript{norm} outperforms Linear FT by a large margin, 75.8 vs. 57.8 on average (see Table \ref{tab: average over all tasks}). It means that only tuning the normalization layers is an efficient method\footnote{According to our observation, the tuning of the normalization layer becomes less important for recent infused fine-tuning methods. Pfeiffer Adapter and MAM Adapter obtain similar results w/o tuning normalization layers.}. With an increasing number of samples, the gap between Full FT and Linear FT\textsubscript{norm} normally becomes narrower, which shows the fine-tuning of high-resource tasks requires less trainable parameters than low-resource.

From the perspective of task variety, we want to make sure all task types appear in our VariousGLUE. COPA, WSC, STS-B and WiC are chosen because of their uniqueness. They are sentence-level ranking, word-level ranking, regression and word-level classification tasks, respectively. The rest are all sentence-level classification tasks. From the perspective of complexity and variety in resources, we choose MNLI and RTE, since MNLI has the biggest number of samples and RTE is the most complex and a low-resource task (see right subplot of Figure \ref{fig: task selection}).

\begin{table*}
\centering
\begin{tabular}{lcc}
    \toprule
    \textbf{Method} & \textbf{\#Tuned} & \textbf{\#Stored} \\
    \hline
    Full FT & $(V+2+n)d + (12d^2 + 13d)L$ &  $(V+2+n)d + (12d^2 + 13d)L$ \\
    Linear FT\textsubscript{norm} & $2d + 4dL$ & $2d + 4dL$ \\
    BitFit & $d + 11dL$ & $d + 11dL$ \\
    Adapter & $(4dr + 2r + 6d)L$ & $(4dr + 2r + 6d)L$ \\
    Pfeiffer Adapter & $(2dr+r+d)L$ & $(2dr+r+d)L$ \\
    LoRA & $4drL$ & $4drL$ \\
    \textbf{Prefix Tuning} & $ld + dm +m + (2md+2d)L$ & $2ldL$ \\
    \textbf{MAM Adapter} & $ld + dm +m + (2dr+r+3d+2md)L$ & $(2dr+r+d+2ld)L$ \\
    \textbf{HiWi for Bias} & $(18dr +3r+5d)L$ & $5dL$ \\
    HiWi for Weight & $(18dr +3r+5d)L$ & $(18dr +3r+5d)L$ \\
    \bottomrule
\end{tabular}
\caption{\label{tab: parameters and storage}
The calculation of trainable and stored parameters. The methods with different calculations for training and storage are highlighted. $V$: the vocabulary size. $n$: the maximum sequence length. $d$: the hidden dimension. $L$: the number of layers. $r$: the bottleneck dimension of the adapter. $m$: the bottleneck dimension of the adapter for Prefix Tuning. $l$: the length of the prefix vector.
}
\end{table*}

\begin{table*}
\centering
\begin{tabular}{lccc}
    \toprule
    \textbf{Method} & \textbf{Hyper-parameter} &\textbf{\#Tuned} & \textbf{\#Stored} \\
    \hline
    Adapter (Table \ref{tab: all}) & $r=18$ & 0.5\% & 0.5\% \\
    Adapter (Figure \ref{fig: overall} and \ref{fig: scale}) & $r=32$ & 0.9\% & 0.9\% \\
    Adapter (Figure \ref{fig: overall} and \ref{fig: scale}) & $r=128$ & 3.6\% & 3.6\% \\
    \hline
    Pfeiffer Adapter (Table \ref{tab: all}) & $r=36$ & 0.5\% & 0.5\% \\
    Pfeiffer Adapter (Figure \ref{fig: overall} and \ref{fig: scale}) & $r=64$ & 0.9\% & 0.9\% \\
    Pfeiffer Adapter (Figure \ref{fig: overall} and \ref{fig: scale}) & $r=128$ & 3.6\% & 3.6\% \\
    \hline
    LoRA (Table \ref{tab: all}) & $r=18$, $s=2$ & 0.5\% & 0.5\% \\
    Prefix Tuning (Table \ref{tab: all}) & $l=36$, $m=36$ & 0.5\% & 0.5\% \\
    MAM Adapter (Table \ref{tab: all}) & $l=18$, $m=18$, $r=18$, $s=2$ & 0.5\% & 0.5\% \\
    \bottomrule
\end{tabular}
\caption{\label{tab: baseline hyper-parameters}
The hyper-parameter values for baselines. $s$ is the scale value for the residual connection.
}
\end{table*}

\section{Number of Trainable Parameters and Storage}
\label{sec: number of tunable parameters and storage}
We summarize the calculation for the number of trainable parameters and storage requirements in Table \ref{tab: parameters and storage}. We only show the calculation of the encoder-only model here. Notably, the newly added classifier for some tasks is excluded from the calculation, since all methods have this same setting. We also show the hyper-parameter values for all baselines used in this paper in Table \ref{tab: baseline hyper-parameters}.

\textbf{Full FT}: The number of parameters for the token and position embeddings is $Vd + 2d + nd$, where $V$ is the vocabulary size, $d$ is the hidden dimension and $n$ is the maximum sequence length. $2d$ here means the number of parameters from the normalization layer since RoBERTa applies a layer normalization after the embedding. For each encoder layer, RoBERTa has four projection layers for key, value, query and output ($4d^2 + 4d$, $4d$ is the number of parameters for bias), and two feed-forward layers. The first feed-forward layer projects the representation from $d$ to $4d$. And the second one projects it back to $d$. So the number of parameters for these two feed-forward layers is $8d^2 + 5d$ ($5d = 4d + d$ for the bias terms). Each layer also has two normalization layers, including $2\times2d$ parameters. Overall, the size of trainable parameters for Full FT is $(V+2+n)d + (12d^2 + 13d)L$, where $L$ is the number of encoder layers.

\textbf{Linear FT\textsubscript{norm}}: Since we only tune the parameters from the normalization layers, so the number of trainable parameters is $2d + 4dL$.

\textbf{BitFit}: We tune all bias terms from the normalization layers and the linear layers, so the size is $d + 11dL$.

\textbf{Adapter}: We insert two adapter layers to each encoder layer and also tune the normalization layers in the encoder layer, so the size of trainable parameters is $(2(dr + r + rd + d) + 4d)L = (4dr + 2r + 6d)L$, where $r$ is the bottleneck dimension of the adapter.

\textbf{Pfeiffer Adapter}: Pfeiffer Adapter inserts a single adapter to each encoder layer and doesn't tune the normalization layers, so the size is $(dr+r+rd+d)L=(2dr+r+d)L$.

\textbf{LoRA}: LoRA applies two adapters in parallel to the projection layers for query and value, respectively. Then the number of trainable parameters is $2(dr+rd)L=4drL$ (no bias terms for the adapter). 

\textbf{Prefix Tuning}: Prefix Tuning applies a re-parameterization trick to expand the number of trainable parameters. Firstly, it defines an embedding in the size of $l\times d$, where $l$ is the length of the prefix vectors. Then this embedding is fed into a large adapter from $d$ to $2dL$. After the adapter, the embedding is in the size of $l \times 2dL$. We then reshape this matrix to $L \times 2 \times ld$, with one prefix vector in the size of $ld$ for the key and another one for the value for each layer. So the number of trainable parameters is $ld + dm +m + 2mdL + 2dL=ld + dm +m + (2md+2d)L$, where $m$ is the bottleneck dimension for the adapter. However, it is not necessary for us to save all these parameters. We can compute the prefix vectors after training and throw away the embedding and adapter, so the size of stored parameters is $2ldL$. 

\textbf{MAM Adapter}: MAM adapter applies Prefix Tuning to the attention module and an adapter in parallel to the MLP module. The number of trainable parameters is the same as the sum of the one for Pfeiffer Adapter and the one for Prefix Tuning, which is $ld + dm +m + (2dr+r+3d+2md)L$. Similar to Prefix Tuning, we can throw away the embedding and adapter for Prefix Tuning after training. So the stored size is $(2dr+r+d+2ld)L$. 

\textbf{HiWi for Bias}: HiWi applies one adapter to the bias term (in the size of $4d$) of the first feed-forward layer and another adapter to the bias term (in the size of $d$) of the second feed-forward layer. To avoid allocating too many trainable parameters to the first adapter, we set the bottleneck dimension for the first adapter as $2r$, and the one for the second adapter as $r$. So the size of trainable parameters is $((8dr + 2r + 8dr + 4d) + (dr + r + dr + d))L=(18dr +3r+5d)L$. After training, we compute the new bias and throw away all adapter parameters. So the stored size is $5dL$.

\textbf{HiWi for Weight} HiWi applies one adapter to the weight (in the size of $4d\times d$) of the first feed-forward layer and another adapter to the weight (in the size of $d\times4d$) of the second feed-forward layer. We only need to swap the order of the above adapters for the bias terms. So the number of trainable parameters stays the same as above, i.e. $(18dr +3r+5d)L$. However, we don't compute the new weight and throw away the adapters for this case, since the size of the weight matrix is larger than the adapter size. So the stored size is still$(18dr +3r+5d)L$.


\end{document}